\documentclass[10pt,twocolumn,letterpaper]{article}

\usepackage[pagenumbers]{iccv} 

\usepackage{booktabs}
\usepackage{colortbl}
\usepackage{multirow}
\usepackage{colortbl}
\usepackage{xcolor}
\usepackage{graphicx}
\usepackage{makecell}
\usepackage{float}

\definecolor{darkgreen}{RGB}{0, 150, 0}
\definecolor{lightgreen}{RGB}{255,250,220}

\definecolor{iccvblue}{rgb}{0.21,0.49,0.74}
\usepackage[pagebackref,breaklinks,colorlinks,allcolors=iccvblue]{hyperref}

\def\confName{ICCV}
\def\confYear{2025}

\title{Boosting Domain Generalized and Adaptive Detection with Diffusion Models: Fitness, Generalization, and Transferability}

\author{
Boyong He$^{1}$\footnotemark[1] ,
Yuxiang Ji$^{1}$\footnotemark[1] ,
Zhuoyue Tan$^{1}$\footnotemark[1] , 
Liaoni Wu$^{1, 2}$\footnotemark[2] ,
\\ 
    $^1$Institute of Artificial Intelligence, Xiamen University \\
    $^2$School of Aerospace Engineering, Xiamen University \\
    \small \texttt{\{boyonghe, yuxiangji, tanzhuoyue\}@stu.xmu.edu.cn}\\
    \small\texttt{ wuliaoni@xmu.edu.cn}\footnotemark[2]
  }

\begin{document}
\maketitle

\renewcommand{\thefootnote}{\fnsymbol{footnote}}
\footnotetext[1]{Equal contribution.}, \footnotetext[2]{Corresponding author.}

\begin{abstract}
Detectors often suffer from performance drop due to domain gap between training and testing data. Recent methods explore diffusion models applied to domain generalization (DG) and adaptation (DA) tasks, but still struggle with large inference costs and have not yet fully leveraged the capabilities of diffusion models. We propose to tackle these problems by extracting intermediate features from a single-step diffusion process, improving feature collection and fusion to reduce inference time by \textbf{75\%} while enhancing performance on source domains (i.e., \textbf{Fitness}). Then, we construct an object-centered auxiliary branch by applying box-masked images with class prompts to extract robust and domain-invariant features that focus on object. We also apply consistency loss to align the auxiliary and ordinary branch, balancing fitness and generalization while preventing overfitting and improving performance on target domains (i.e., \textbf{Generalization}). Furthermore, within a unified framework, standard detectors are guided by diffusion detectors through feature-level and object-level alignment on source domains (for DG) and unlabeled target domains (for DA), thereby improving cross-domain detection performance (i.e., \textbf{Transferability}). Our method achieves competitive results on 3 DA benchmarks and 5 DG benchmarks. Additionally, experiments on COCO generalization benchmark demonstrate that our method maintains significant advantages and show remarkable efficiency in large domain shifts and low-data scenarios. Our work shows the superiority of applying diffusion models to domain generalized and adaptive detection tasks and offers valuable insights for visual perception tasks across diverse domains. The code is available at \href{https://github.com/heboyong/Fitness-Generalization-Transferability}{Fitness-Generalization-Transferability}.
\end{abstract}

\section{Introduction}

Distribution discrepancies between training data and real-world environments are inevitable in practical applications of object detection. Factors such as weather variations, sensor differences, diverse lighting conditions, and environmental noise contribute to domain shift between training and testing datasets. This domain gap causes even the state-of-the-art (SOTA) detectors~\cite{faster_rcnn,yolov3,fcos,dino_detector} to suffer from significant performance degradation when applied to diverse and unseen domains~\cite{dg, dg_survey}. 

To address domain gaps in object detection, researchers have developed two main approaches: Domain Adaptation (DA) and Domain Generalization (DG). DA leverages unlabeled target domain data through feature alignment~\cite{chen2018da-faster,saito2019swda,AT}, style transfer~\cite{gan,cyclegan,lods}, and self-training with pseudo-labels~\cite{htcn,cao2023cmt,MTTrans}. However, since target data is often unavailable beforehand, DG methods focus on data augmentation~\cite{mixstyle,stylemix}, domain-invariant feature extraction~\cite{oamix,Diversification}, and adversarial training~\cite{dg_adver,localized} to build robust detectors using only source domain data that perform well across unseen domains.

Vision foundation models offer fresh perspectives on domain challenges. DDT~\cite{ddt} and GDD~\cite{gdd} utilize diffusion models~\cite{ddpm,song2020ddim,sd3} to build robust detectors for DA and DG tasks, outperforming previous methods. However, these approaches suffer from computational inefficiency due to large parameters and multi-step denoising processes, limiting their practical deployment. Moreover, existing frameworks fail to fully exploit the multimodal capabilities of diffusion models or develop specialized architectures for DG and DA tasks, indicating untapped potential for improving generalization and adaptation performance.

In this paper, we address these limitations through three key strategies. First, unlike DDT~\cite{ddt} and GDD~\cite{gdd} which rely on computationally expensive multi-step diffusion processes, we implement single-step feature extraction with optimized collection and fusion structures. Second, we design an object-centered auxiliary branch using box-masked images and class prompts, leveraging diffusion models' multi-modal capabilities. Finally, we align regular and auxiliary branch through consistency loss, enabling domain-invariant feature learning without affecting inference speed. Compared to SOTA method GDD~\cite{gdd}, our approach reduces inference time by \textbf{75\%} while achieving \textbf{\{2.7, 0.6, 3.8, 4.8, 3.3}\}\% mAP improvements across 5 DG benchmarks.

We also explore transferring the powerful generalization capabilities of diffusion detectors to standard detectors. Building upon the DDT~\cite{ddt} and GDD~\cite{gdd} approaches, we propose a unified transfer framework that aligns at both feature and object levels, applicable to both DA and DG tasks. Benefiting from our improved diffusion detector, diffusion-guided detectors achieve \textbf{\{0.4, 1.6, 0.8, 1.8, 1.6}\}\% mAP improvements across 5 DG benchmarks compared to GDD~\cite{gdd}, and \textbf{\{7.9, 6.6, 1.7}\}\% improvements on 3 DA benchmarks compared to DDT~\cite{ddt}.

Furthermore, we evaluate diffusion detectors on larger-scale benchmarks, training on different proportions (1\% and 100\%) of COCO~\cite{coco} and testing across 11 cross-domain datasets. Compared to advanced architectures (ConvNeXt~\cite{Convnext}, Swin~\cite{swin}, VIT~\cite{vit}) and pre-trained models (GLIP~\cite{glip}), our method shows significant advantages, particularly in scenarios with large domain shifts or limited data. Our results on multiple DG and DA benchmarks demonstrate our approach offers an efficient solution to addressing domain gap challenges in detection tasks.

\section{Related Work}

\subsection{DG and DA for Object Detection}

Detectors suffer performance drop when deployed in environments different from training data. DA approaches address this by aligning feature distributions through adversarial learning~\cite{chen2018da-faster,saito2019swda,crda,AT} or consistency-based learning with pseudo-labels~\cite{htcn,AT,deng2023HT,cao2023cmt}. However, these methods require target domain data during training. To overcome this limitation, DG methods develop robust models without accessing target data, primarily through data augmentation~\cite{mixstyle,stylemix,randaugment,Diversification}, adversarial training~\cite{dg_adver,localized}, meta-learning~\cite{metareg,vibdg}, and style transfer~\cite{gan,cyclegan}. Recent works extend these to detection tasks via multi-view learning~\cite{zhou2022mga}, feature disentanglement~\cite{gnas}, augmentation~\cite{oamix,Diversification} and causal inference~\cite{ufr}. Despite these advances, current methods still struggle to effectively handle the complex domain shifts present in real-world environments.

\subsection{Diffusion-Based Applications}
Diffusion models~\cite{ho2020ddpm,song2020ddim,sd1,dalle,imagen,sd3} have demonstrated exceptional capabilities in image generation and representation learning~\cite{ddpmseg,xu2023odise}. Their noise-adding and denoising mechanism provide natural robustness against visual perturbations~\cite{hyperfeatures,tang2023emergent}, making them promising for domain generalization tasks. Recent works have successfully applied diffusion-derived features to semantic segmentation~\cite{ddpmseg}, panoptic segmentation~\cite{xu2023odise}, and image correspondence~\cite{tang2023emergent,hyperfeatures}. Specifically, DDT~\cite{ddt} and GDD~\cite{gdd} explored diffusion models for DA and DG detection task respectively, achieving significant improvements that demonstrate the potential of building robust diffusion-based detectors. Our approach builds upon the work~\cite{ddt,gdd}, substantially optimizing inference efficiency while enhancing both fitness and generalization capabilities, showing strong performance across complex cross-domain scenarios and various data scales.

\section{Method}
\subsection{Preliminaries}
\noindent \textbf{Object Detection:} Object detection aims to locate and classify objects within images. Given an input image $I$, a detector outputs bounding boxes $\mathcal{B} = \{b_i\}_{i=1}^{N}$ with corresponding class labels $\mathcal{C} = \{c_i\}_{i=1}^{N}$, where $c_i \in \{1,2,...,K\}$ for $K$ categories. We adopt Faster R-CNN~\cite{faster_rcnn} as our default detector, which first generates region proposals via a Region Proposal Network (RPN), then uses Region of Interest (ROI) head to extract features for classification and bounding box regression.

\noindent \textbf{DG and DA for Object Detection:} Domain Generalization (DG) and Domain Adaptation (DA) address distributional shift between domains. With source domain $\mathcal{D}_S = \{(x_i^S, y_i^S)\}_{i=1}^{N_S}$ and target domain $\mathcal{D}_T = \{x_j^T\}_{j=1}^{N_T}$, the goal is optimal performance on $\mathcal{D}_T$. DG works without accessing target data during training, while DA leverages unlabeled target data to adapt the model.

\noindent \textbf{Diffusion Process of Diffusion Models:} Diffusion models define a forward process that gradually adds Gaussian noise to data samples. This forward process transforms a data sample $\mathbf{x}_0$ into noise $\mathbf{x}_T$ through a Markov chain. The transition probability is given by: $q(\mathbf{x}_t|\mathbf{x}_{t-1}) = \mathcal{N}(\mathbf{x}_t; \sqrt{1-\beta_t}\mathbf{x}_{t-1}, \beta_t\mathbf{I})$, where $\beta_t \in (0,1)$ controls the noise added at each step. Remarkably, we can sample $\mathbf{x}_t$ directly from $\mathbf{x}_0$ using: $\mathbf{x}_t = \sqrt{\bar{\alpha}_t}\mathbf{x}_0 + \sqrt{1-\bar{\alpha}_t}\mathbf{\epsilon}$, where $\bar{\alpha}_t = \prod_{i=1}^{t}(1-\beta_i)$ accumulates the noise schedule effects and $\mathbf{\epsilon} \sim \mathcal{N}(\mathbf{0}, \mathbf{I})$ is standard Gaussian noise. 

\subsection{Feature Collection and Fusion}

\begin{figure}[t]
    \centering
    \includegraphics[width=0.48\textwidth]{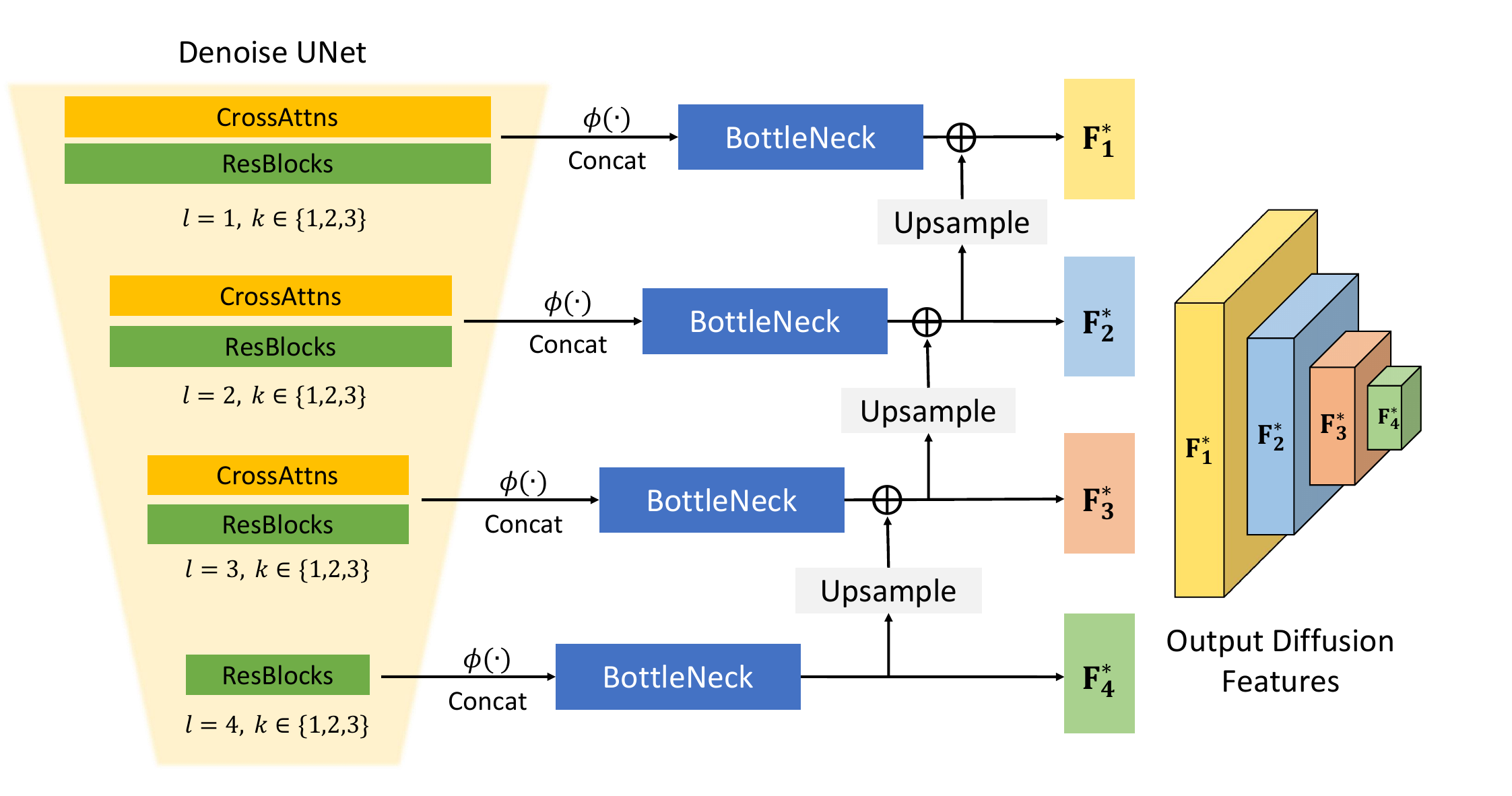}
    \caption{Feature collection and fusion from UNet on single-step diffusion process .}
    \label{fig:feature extractor}
\end{figure}

\noindent\textbf{{Collection}}: 
To optimize inference efficiency, we propose to extract rich and robust features from a single-step diffusion process rather than across multiple steps in DDT~\cite{ddt}. Given an input image $\mathbf{x}_0$, we apply the forward diffusion process to obtain a noisy sample $\mathbf{x}_t$ at a specific timestep $t$. From the noise predictor $\mathcal{F}_{\theta}$, we extract two comprehensive sets of features: 12 feature groups from ResNet blocks in the UNet upsampling structure, denoted as $\mathbf{s}_{res}^{l,k} \in \mathbb{R}^{C_{l,k}^{res} \times H_{l,k}^{res} \times W_{l,k}^{res}}$, where $l \in \{1,2,3,4\}$ indicates the layer and $k \in \{1,2,3\}$ specifies the block position within each layer; and 9 feature groups from cross-attention blocks, denoted as $\mathbf{s}_{att}^{l,k} \in \mathbb{R}^{C_{l,k}^{att} \times H_{l,k}^{att} \times W_{l,k}^{att}}$, where $l \in \{1,2,3\}$ and $k \in \{1,2,3\}$. This focused extraction strategy captures the multi-scale semantic information while maintaining computational efficiency as shown in the left part of Fig.~\ref{fig:feature extractor}.

\noindent\textbf{{Fusion}}: Our feature fusion approach leverages the hierarchical structure as shown in Fig.~\ref{fig:feature extractor}. From upsampling stages, we obtain features at four different spatial resolutions $\mathbf{F}_l$ for $l \in \{1,2,3,4\}$. Different from processing each layer separately as in GDD~\cite{gdd}, we concatenate all features of the same scale into a unified representation: $\mathbf{F}_l^{\oplus} = \phi(\mathbf{s}_{res}^{l,1}, \mathbf{s}_{res}^{l,2}, \mathbf{s}_{res}^{l,3}, \mathbf{s}_{att}^{l,1}, \mathbf{s}_{att}^{l,2}, \mathbf{s}_{att}^{l,3})$ for each layer $l$, where $\phi(\cdot)$ denotes channel-wise concatenation. Each concatenated feature group then undergoes a single bottleneck projection: $\mathbf{F}_l^{p} = \mathcal{B}(\mathbf{F}_l^{\oplus})$, where $\mathcal{B}(\cdot)$ represents the bottleneck operation that projects features to dimensions $C_l = 256 \times 2^{l-1}$. Then we implement skip connections by adding features of matching resolutions: $\mathbf{F}_{l-1}^{*} = \mathcal{U}(\mathbf{F}_l^{p}) + \mathbf{F}_{l-1}^{p}$, where $\mathcal{U}(\cdot)$ denotes the upsampling operation, preserving fine-grained details that might otherwise be lost with direct upsampling in GDD~\cite{gdd}. The final feature pyramid $\mathbf{F}^{final} = \{\mathbf{F}_l^{*}\}_{l=1}^4$ with $\mathbf{F}_l^{*} \in \mathbb{R}^{C_l \times H/2^{l+1} \times W/2^{l+1}}$ (where $C_l = 256 \times 2^{l-1}$ for $l \in \{1,2,3,4\}$) aligns structurally with standard ResNet~\cite{resnet} outputs, ensuring compatibility with detection heads while effectively compensating for the performance limitations of single-step feature extraction.

\subsection{Dual-branch of Diff. Detector} 

\begin{figure*}[ht]
    \centering
    \begin{minipage}{0.68\textwidth}
        \centering
        \includegraphics[width=\textwidth]{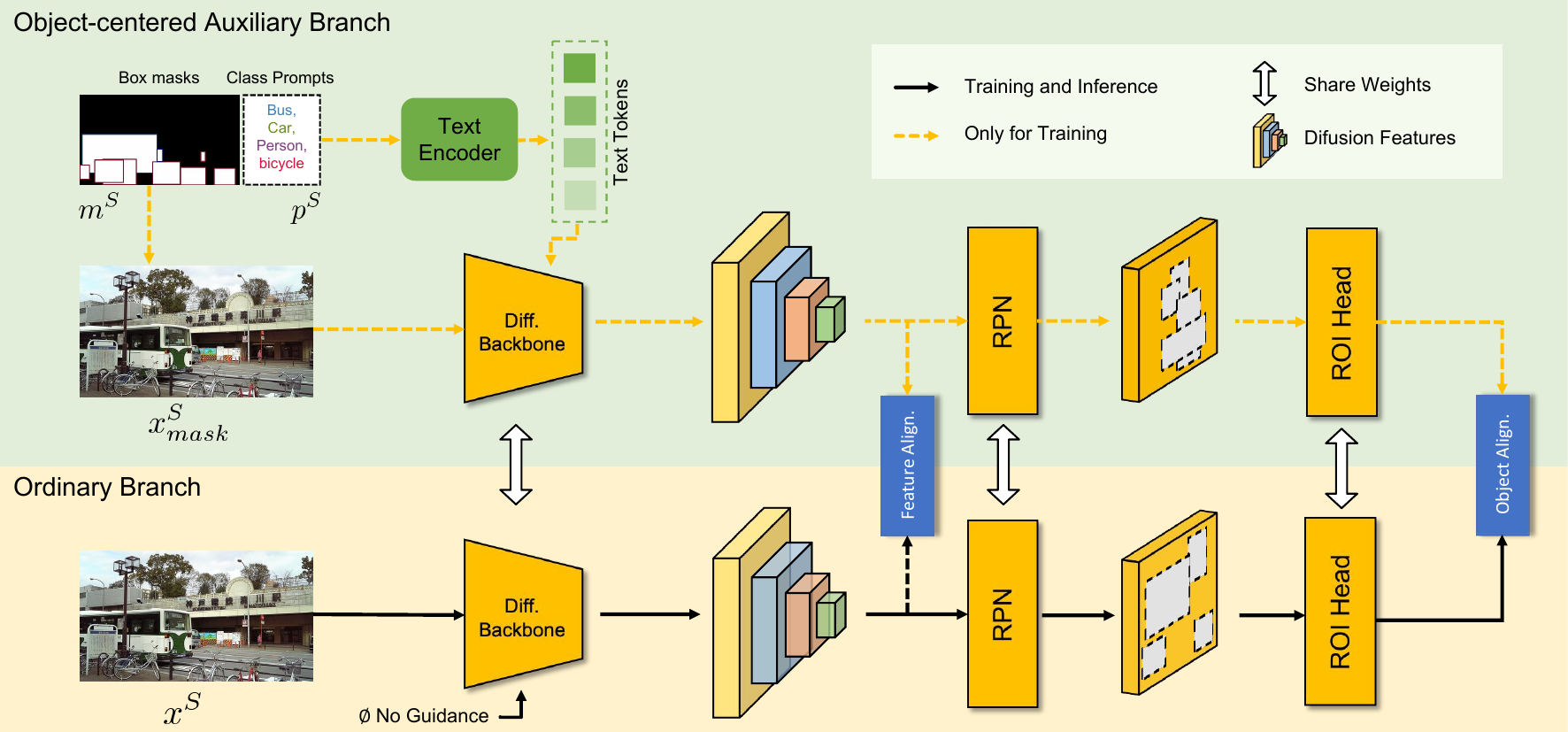}
        \caption{Our proposed dual-branch detection framework: Object-centered Auxiliary Branch (\textbf{top}) and Ordinary Detection Branch (\textbf{bottom}), unified through novel feature-level and object-level consistency alignments.}
        \label{fig:frameworks}
    \end{minipage}
    \hfill
    \begin{minipage}{0.31\textwidth}
        \centering
        \includegraphics[width=\textwidth]{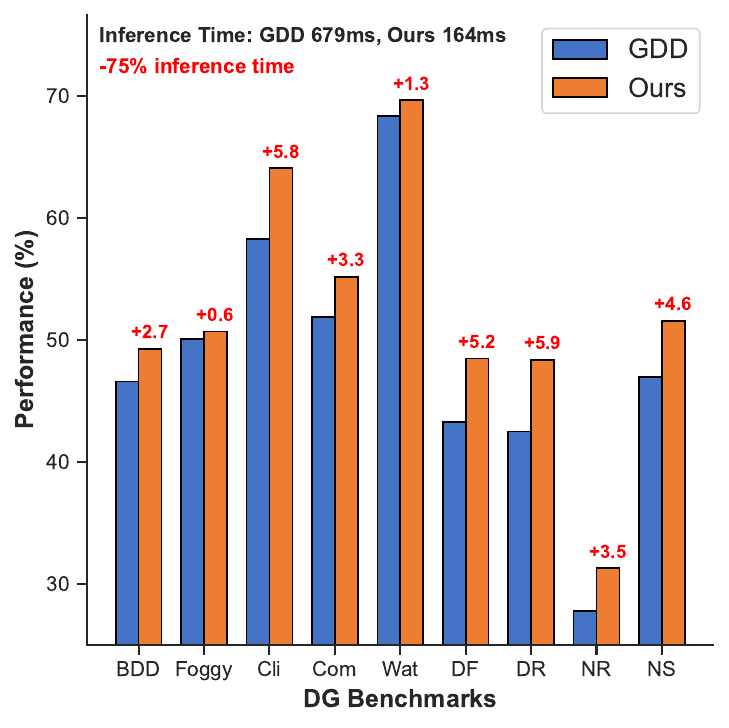}
        \caption{Performance comparison with GDD~\cite{gdd} across DG benchmarks. Our method shows improvements with 75\% less inference time.}
        \label{fig:comparisons}
    \end{minipage}
\end{figure*}

\noindent\textbf{Ordinary Branch:} Our diffusion backbone extracts features as described above to construct the \textit{Diff. Detector} ($\mathcal{F}_{\textnormal{diff}}$). Given source domain images ${x}^S$ and labels ${y}^S$, we process inputs through the frozen diffusion structure and trainable feature extraction components to generate the feature pyramid $\mathbf{F}_{ord}^{final}$. These features feed into RPN and ROI heads, the training objective is:

\begin{align}
    \mathcal{L}_{ord} = \mathcal{L}(\mathcal{F}_{\textnormal{diff}}({x}^S), {y}^S)
\end{align}
where $\mathcal{L}$ combines the classification, bounding box regression, and region proposal losses of Faster R-CNN, as illustrated in the bottom panel of Fig.~\ref{fig:frameworks}.

\noindent\textbf{Object-centered Auxiliary Branch:} Diffusion models demonstrate powerful multimodal understanding capabilities through text-conditioned image generation, yet this advantage remains unexplored in DDT~\cite{ddt} and GDD~\cite{gdd}. We propose an object-centered auxiliary branch that leverages the cross-modal capabilities of diffusion to enhance feature representation and generalization for detection.

Specifically, our Object-centered Auxiliary Branch takes source domain images $x^S$ along with their corresponding bounding box masks $m^S$ and class prompts $p^S$ as inputs. First, we generate object-centered images via $x_{mask}^S = x^S \odot m^S$, where $\odot$ represents element-wise multiplication and $m^S$ is a binary mask derived from ground-truth bounding boxes, with non-object regions set to zero. We then feed $x_{mask}^S$ and class prompts $p^S$ into the diffusion model's conditioning process, where $\mathbf{z} = \mathcal{E}(x_{mask}^S)$ is the latent representation from the image encoder and $\mathbf{c} = \mathcal{T}(p^S)$ is the semantic embedding from the text encoder. The cross-attention mechanism facilitates interaction between textual conditions and image features, focusing attention on regions relevant to the prompted class.

This process enables the diffusion model to generate features with enhanced focus on specified object categories. We extract features $\mathbf{F}_{aux}^{final}$ from the same positions as in the ordinary branch and apply identical detection heads as shown in the top part of Fig.~\ref{fig:frameworks}. The objective function for this auxiliary branch is:

\begin{align}
    \mathcal{L}_{aux} = \mathcal{L}(\mathcal{F}_{\textrm{diff}}(x_{mask}^S, p^S), y^S)
\end{align}

By fully leveraging ground-truth labels during training, this branch provides additional supervision that better exploits the multimodal capabilities of diffusion models, leading to more domain-invariant and object-centered representations.

\subsection{Dual-branch Consistency Loss}
To promote stronger domain-invariant feature learning in the ordinary branch, we align its features and ROI outputs with those from the Object-centered branch. This alignment enables the inference-time model, which relies solely on the ordinary branch, to benefit from domain-invariant features, thereby enhancing generalization on unseen domains. Specifically, for features $\mathbf{F}_{ord}^{final}$ and $\mathbf{F}_{aux}^{final}$ from the two branches, we align them using mean squared error loss:

\begin{align}
\mathcal{L}_{feature} &= \|\mathbf{F}_{ord}^{final} - \mathbf{F}_{aux}^{final}\|_2^2 
\end{align}

Furthermore, referencing CrossKD~\cite{crosskd}, we design cross-head alignment to align the ROI outputs from both branches, including bounding box alignment and class alignment. The ROI outputs consist of $\mathbf{B}_{ord}$ and $\mathbf{B}_{aux}$ for bounding box predictions, and $\mathbf{C}_{ord}$ and $\mathbf{C}_{aux}$ for classification logits from the ordinary and auxiliary branches respectively:

\begin{align}
\mathcal{L}_{box} &= |\mathbf{B}_{ord} - \mathbf{B}_{aux}| \\
\mathcal{L}_{cat} &= \sum_{i} P_{ord}^{\tau}(i) \log\frac{P_{ord}^{\tau}(i)}{P_{aux}^{\tau}(i)}
\end{align}
where $P_{ord}^{\tau}$ and $P_{aux}^{\tau}$ represent the softened probability distributions obtained by applying softmax with temperature $\tau$ to the classification logits $\mathbf{C}_{ord}$ and $\mathbf{C}_{aux}$ respectively.

The full alignment objective is formulated as a weighted combination of feature and output consistency:
\begin{align}
\label{l_cons}
\mathcal{L}_{cons} = \mathcal{L}_{feature} + \gamma \cdot (\mathcal{L}_{box} + \mathcal{L}_{cat})
\end{align}
where $\gamma$ is a weighting parameter that balances the importance between feature-level and output-level alignments.

\subsection{Full Objective}
Combining the ordinary branch detection loss, auxiliary branch detection loss, and consistency loss, the full training objective of our diff. detector is formulated as:
\begin{align}
\label{l_total}
\mathcal{L}_{total} = \mathcal{L}_{ord} + \mathcal{L}_{aux} + \lambda\mathcal{L}_{cons}
\end{align}
where $\lambda$ is a weighting parameter that controls the contribution of the consistency regularization between the two branches.

\subsection{Dual-branch and Consistency Loss for Generalization}

From a representation learning perspective, achieving domain generalization requires disentangling task-relevant domain-invariant features from domain-specific features. We can formally decompose image features as:
\begin{equation}
\Phi(x) = \alpha \Phi_{inv}(x) + \Phi_{spe}(x)
\end{equation}
where $\Phi_{inv}(x)$ represents domain-invariant features essential for detection tasks (e.g., object geometry, semantic attributes), $\Phi_{spe}(x)$ represents domain-specific features (e.g., lighting conditions, background environments), and $\alpha$ is a coefficient indicating the contribution weight of invariant features.

\noindent\textbf{Dual-branch Mechanism for Feature Disentanglement:} Our dual-branch architecture implements this disentanglement principle through complementary learning paths. The Ordinary Branch processes complete images, initially capturing both $\Phi_{inv}(x)$ and $\Phi_{spe}(x)$ with a smaller $\alpha$, while the Object-centered Branch emphasizes $\Phi_{inv}(x)$ with a larger $\alpha$ while suppressing $\Phi_{spe}(x)$ through object masks $m^S$ and class prompts $p^S$, such that $\Phi_{aux}(x_{mask}^S, p^S)$ primarily contains $\Phi_{inv}(x)$ with enhanced $\alpha$.

\noindent\textbf{Consistency Loss as Knowledge Distillation:} The consistency loss functions as a knowledge transfer mechanism that guides the ordinary branch toward domain-invariant representations. By aligning features and detection outputs between branches, we guide the ordinary branch to amplify its domain-invariant features by increasing $\alpha$:

\begin{equation}
\mathcal{F}_{\textnormal{diff}}(x^S) \xrightarrow{\mathcal{L}_{cons}} \mathcal{F}_{\textnormal{diff}}(x_{mask}^S, p^S)
\end{equation}

From a risk minimization perspective, minimizing these consistency losses helps reduce the domain generalization error bound:
\begin{equation}
\mathcal{R}_{T}(h) \leq \mathcal{R}_{S}(h) + d_{\mathcal{H}}(\mathcal{D}_{S}, \mathcal{D}_{T}) + \delta
\end{equation}

The consistency loss reduces $d_{\mathcal{H}}(\mathcal{D}_{S}, \mathcal{D}_{T})$ by encouraging features with domain-invariant components, ensuring predictions become invariant across domains $d_A$ and $d_B$:
\begin{equation}
    \min_{\mathcal{F}_{\textnormal{diff}}} \left\| \mathbb{E}[Y|\mathcal{F}_{\textnormal{diff}}(X), D=d_A] - \mathbb{E}[Y|\mathcal{F}_{\textnormal{diff}}(X), D=d_B] \right\|
\end{equation}

Additionally, this loss serves as a regularization term that mitigates overfitting on source domains:
\begin{equation}
\min_{\mathcal{F}_{\textnormal{diff}} \in \mathcal{F}} \mathcal{L}_{task}(\mathcal{F}_{\textnormal{diff}}(x^S)) + \lambda \cdot \mathcal{L}_{cons}(\mathcal{F}_{\textnormal{diff}}(x^S), \mathcal{F}_{\textnormal{diff}}(x_{mask}^S, p^S))
\end{equation}

This constraint effectively shrinks the hypothesis space, with both branches sharing the same feature extractor of $\mathcal{F}_{\textnormal{diff}}$ but operating on different inputs.


\subsection{Unified Transfer Framework for DG and DA}

\begin{figure}[!ht]
    \centering
    \includegraphics[width=1.0\columnwidth]{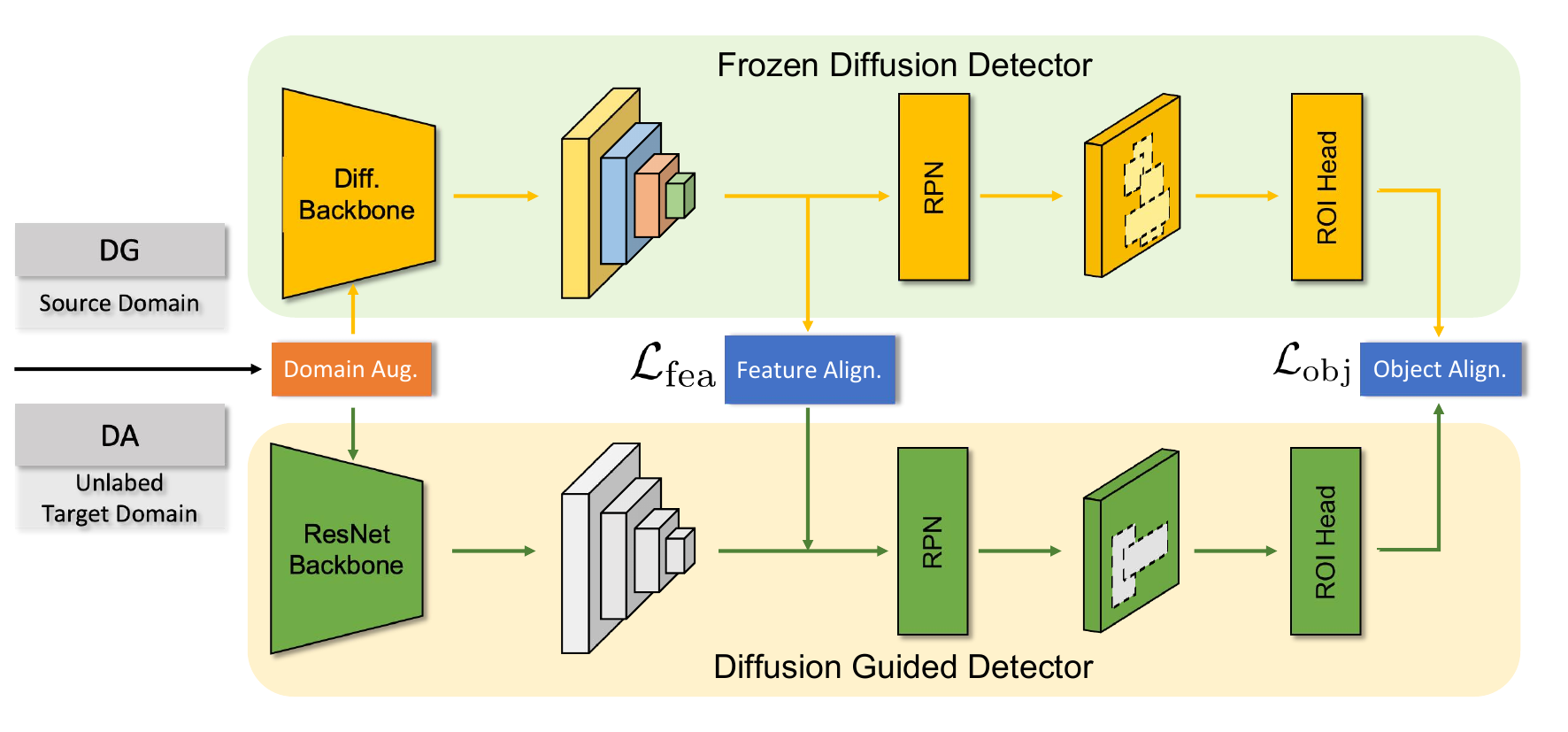}
    \caption{Unified transfer framework for DG and DA with feature- and object-level alignment.}
    \label{fig:transfer}
\end{figure}

Given the strong generalization capabilities demonstrated by the diffusion detector, we aim to transfer these capabilities to ordinary detectors as shown in Fig.~\ref{fig:transfer}. We present a unified transfer framework that consolidates approaches from both DDT~\cite{ddt} and GDD~\cite{gdd} and follow their settings, establishing alignment at feature ($\mathcal{L}_{\text{fea}}$) and object levels ($\mathcal{L}_{\text{obj}}$), as:
\begin{equation}
\mathcal{L}_{\text{transfer}} = \mathcal{L}_{\text{det}} + \mathcal{L}_{\text{fea}} + \mathcal{L}_{\text{obj}}
\end{equation}
where $\mathcal{L}_{\text{det}}$ is detection loss from supervised learning on source domain.

This framework adapts to each task differently. For DA, it generates pseudo-labels in the target domain for object-level alignment while aligning representations for feature-level alignment. For DG, both alignments occur only within source domains, creating a knowledge distillation process that transfers domain-invariant qualities from diffusion to ordinary detector.


\section{Experiments}
\subsection{DG and DA Benchmarks}

~~~~\textbf{(1) Cross Camera}: Cityscapes~\cite{cityscapes} (2,975 training images from 50 cities) to BDD100K~\cite{bdd100k} day-clear split with 7 shared categories following SWDA~\cite{saito2019swda}.

\textbf{(2) Adverse Weather}: Cityscapes to FoggyCityscapes~\cite{foggy} with the most challenging 0.02 split to evaluate robustness under degraded visibility conditions.

\textbf{(3) Real to Artistic}: VOC~\cite{voc} (16,551 real-world images) to Clipart (1K images, 20 categories), Comic (2K images, 6 categories), and Watercolor (2K images, 6 categories)~\cite{clipart_comic_watercolor} following AT~\cite{AT}.

\textbf{(4) Diverse Weather Datasets}: Daytime-Sunny (26,518 images) to Night-Sunny (26,158 images), Night-Rainy (2,494 images), Dusk-Rainy (3,501 images), and Daytime-Foggy (3,775 images) following~\cite{cdsd} and~\cite{oamix}.

\textbf{(5) Corruption Benchmark}: Cityscapes-C~\cite{cityscapes-c} with 15 corruption types (noise, blur, weather, and digital perturbations) at 5 severity levels following OADG~\cite{oamix}.

We test our DG and DA methods on benchmarks \textbf{(1)-(5)} and \textbf{(1)-(3)} respectively, while comparing with existing works. Additionally, we propose a larger-scale DG benchmark: \textbf{COCO Generalization Benchmark}, trained on COCO2017~\cite{coco} dataset and tested on 11 various datasets as shown in Tab.~\ref{tab:coco}.

\subsection{Implementation Details}
\noindent \textbf{Experimental Settings}: Our experimental settings generally align with DDT~\cite{ddt} and GDD~\cite{gdd}. Specifically, all experiments are implemented on MMdetection~\cite{mmdetection}. We consistently use SGD optimizer with a learning rate of 0.02 and a total batch size of 16, training for 20K iterations on two 4090 GPUs. For diff. detector (\textit{\small\textbf{Diff. Detector}}), we apply frozen official weights from Stable Diffusion v1.5 (\textit{\small\textbf{SD-1.5}}) and v2.1 (\textit{\small\textbf{SD-2.1}}) provided by StabilityAI. In our diffusion guided DG and DA experiments, we employ Faster R-CNN~\cite{faster_rcnn} with R101~\cite{resnet} as our baseline detector (\textit{\small\textbf{Diff. Guided, R101}}). We report $\text{AP}_{50}$ for each category and mAP across all categories. For Cityscapes-C~\cite{cityscapes-c}, we report mPC (average $\text{AP}_{50:95}$ across 15 corruptions with 5 levels) following~\cite{oamix} as shown in Tab.~\ref{tab:cityscapes-c}. 

\noindent \textbf{Data Augmentations}: We follow GDD's~\cite{gdd} domain augmentation strategy (\textit{Domain Aug.}) in all experiments, using image-level (color and spatial transformations) and domain-level augmentations (\textit{FDA}~\cite{fda}, \textit{Histogram Matching}, and \textit{Pixel Distribution Matching}). For DG experiments, we apply these augmentations only on source domains, while for DA experiments, we implement them between source and target domains.
\textbf{\textit{{Code and more detailed settings are provided in supplementary materials, along with additional experimental analyses, class-wise results, and qualitative visualizations.}}}

\subsection{Results and Comparisons}

\begin{table}[ht]
    \centering
    \caption{DG and DA Results (\%) on BDD100K.}
    \vspace{-4pt}
    \label{tab:bdd100k}
    \setlength{\tabcolsep}{3pt}
    \resizebox{\columnwidth}{!}{%
    \begin{tabular}{l|ccccccc>{\columncolor{gray!20}}c}
    \toprule
    \textbf{Methods} & \textbf{Bike} & \textbf{Bus} & \textbf{Car} & \textbf{Motor} & \textbf{Psn.} & \textbf{Rider} & \textbf{Truck} & \textbf{mAP} \\
    \midrule
    \multicolumn{9}{c}{\textit{\textbf{DG methods} (without target data)}} \\
    CDSD~\cite{cdsd} \textit{\scriptsize{(\textcolor{darkgreen}{CVPR'22})}} & 22.9 & 20.5 & 33.8 & 14.7 & 18.5 & 23.6 & 18.2 & 21.7 \\
    SHADE~\cite{shade} \textit{\scriptsize{(\textcolor{darkgreen}{ECCV'22})}} & 25.1 & 19.0 & 36.8 & 18.4 & 24.1 & 24.9 & 19.8 & 24.0 \\
    MAD~\cite{mad} \textit{\scriptsize{(\textcolor{darkgreen}{CVPR'23})}} & - & - & - & - & - & - & - & 28.0 \\
    SRCD~\cite{srcd} \textit{\scriptsize{(\textcolor{darkgreen}{TNNLS'24})}} & 24.8 & 21.5 & 38.7 & 19.0 & 25.7 & 28.4 & 23.1 & 25.9 \\
    DDT \footnotesize{\textit{(SD-1.5)}}\normalsize~\cite{ddt}  \textit{\scriptsize{(\textcolor{darkgreen}{MM'24})}} & - & - & - & - & - & - & - & 32.7 \\
    GDD \footnotesize{\textit{(SD-1.5)}}\normalsize~\cite{gdd}  \textit{\scriptsize{(\textcolor{darkgreen}{CVPR'25})}} & 38.9 & 31.0 & 71.5 & 37.6 & 61.5 & 47.0 & 38.5 & 46.6 \\
    GDD \footnotesize{\textit{(R101)}}\normalsize~\cite{gdd}  \textit{\scriptsize{(\textcolor{darkgreen}{CVPR'25})}} & 38.4 & 33.4 & 72.0 & \textbf{38.3} & 60.3 & 47.0 & 35.0 & 46.3 \\
    \midrule
    \rowcolor{lightgreen} 
    \textit{\textbf{Ours \footnotesize{(Diff. Detector, SD-1.5)}}} & \textbf{41.2} & \textbf{41.7} & \textbf{72.7} & 37.2 & \textbf{62.8} & \textbf{48.7} & \textbf{40.5} & \textbf{49.3} \\
    \textit{\textbf{Ours \footnotesize{(Diff. Guided, R101)}}} \normalsize & 39.4 & 34.1 & 72.2 & 37.4 & 61.3 & 46.9 & 35.7 & 46.7 \\
    \midrule \midrule
    \multicolumn{9}{c}{\textit{\textbf{DA methods} (with unlabeled target data)}} \\ 
    EPM~\cite{hsu2020epm} \textit{\scriptsize{(\textcolor{darkgreen}{ECCV'20})}} & 20.1 & 19.1 & 55.8 & 14.5 & 39.6 & 26.8 & 18.8 & 27.8 \\
    TDD~\cite{tdd} \textit{\scriptsize{(\textcolor{darkgreen}{CVPR'22})}} & 28.8 & 25.5 & 53.9 & 24.5 & 39.6 & 38.9 & 24.1 & 33.6 \\
    PT~\cite{pt} \textit{\scriptsize{(\textcolor{darkgreen}{ICML'22})}}& 28.8 & 33.8 & 52.7 & 23.0 & 40.5 & 39.9 & 25.8 & 34.9 \\
    SIGMA~\cite{li2022sigma} \textit{\scriptsize{(\textcolor{darkgreen}{CVPR'22})}}& 26.3 & 23.6 & 64.1 & 17.9 & 46.9 & 29.6 & 20.2 & 32.7 \\
    SIGMA++~\cite{li2023sigma++} \textit{\scriptsize{(\textcolor{darkgreen}{TPAMI'23})}}& 27.1 & 26.3 & 65.6 & 17.8 & 47.5 & 30.4 & 21.1 & 33.7 \\
    NSA~\cite{nsa} \textit{\scriptsize{(\textcolor{darkgreen}{ICCV'23})}}& - & - & - & - & - & - & - & 35.5 \\
    HT~\cite{deng2023HT} \textit{\scriptsize{(\textcolor{darkgreen}{CVPR'23})}}& 38.0 & 30.6 & 63.5 & 28.2 & 53.4 & 40.4 & 27.4 & 40.2 \\
    MTM~\cite{mtm} \textit{\scriptsize{(\textcolor{darkgreen}{AAAI'24})}} & 28.0 & 28.8 & 68.8 & 23.8 & 53.7 & 35.1 & 23.0 & 37.3 \\
    CAT~\cite{cat} \textit{\scriptsize{(\textcolor{darkgreen}{CVPR'24})}} & 34.6 & 31.7 & 61.2 & 24.4 & 44.6 & 41.5 & 31.4 & 38.5 \\
    DDT \footnotesize{\textit{(R101)}}\normalsize~\cite{ddt} \textit{\scriptsize{(\textcolor{darkgreen}{MM'24})}} & 40.3 & 32.3 & 66.7 & 31.8 & 59.1 & 41.6 & 31.8 & 43.4 \\
    \midrule
    \rowcolor{lightgreen} 
    \textit{\textbf{Ours \footnotesize{(Diff. Guided, R101)}}} & \textbf{43.6} & \textbf{42.9} & \textbf{75.2} & \textbf{40.5} & \textbf{64.6} & \textbf{49.6} & \textbf{42.9} & \textbf{51.3} \\
    \bottomrule
    \end{tabular}%
    }
    \end{table}

\begin{table}[ht]
    \centering
    \caption{DG and DA Results (\%) on FoggyCityscapes.}
    \vspace{-4pt}
    \label{tab:FoggyCityscapes}
    \setlength{\tabcolsep}{2pt}  
    \resizebox{\columnwidth}{!}{%
    \begin{tabular}{l|cccccccc>{\columncolor{gray!20}}c}
    \toprule
    \textbf{Methods} & \textbf{Bus} & \textbf{Bike} & \textbf{Car} & \textbf{Motor} & \textbf{Psn.} & \textbf{Rider} & \textbf{Train} & \textbf{Truck} & \textbf{mAP} \\
    \midrule
    \multicolumn{10}{c}{\textit{\textbf{DG methods}}} \\
    DIDN~\cite{didn} \textit{\scriptsize{(\textcolor{darkgreen}{CVPR'21})}} & 35.7 & 33.1 & 49.3 & 24.8 & 31.8 & 38.4 & 26.5 & 27.7 & 33.4 \\
    FACT~\cite{fact} \textit{\scriptsize{(\textcolor{darkgreen}{CVPR'21})}} & 27.7 & 31.3 & 35.9 & 23.3 & 26.2 & 41.2 & 3.0 & 13.6 & 25.3 \\
    FSDR~\cite{fsdr} \textit{\scriptsize{(\textcolor{darkgreen}{CVPR'22})}} & 36.6 & 34.1 & 43.3 & 27.1 & 31.2 & 44.4 & 11.9 & 19.3 & 31.0 \\
    MAD~\cite{mad} \textit{\scriptsize{(\textcolor{darkgreen}{CVPR'23})}} & 44.0 & 40.1 & 45.0 & 30.3 & 34.2 & 47.4 & 42.4 & 25.6 & 38.6 \\
    DDT \footnotesize{\textit{(SD-1.5)}}\normalsize~\cite{ddt} \textit{\scriptsize{(\textcolor{darkgreen}{MM'24})}} & - & - & - & - & - & - & - & - & 36.1 \\
    GDD \footnotesize{\textit{(SD-1.5)}}\normalsize~\cite{gdd}  \textit{\scriptsize{(\textcolor{darkgreen}{CVPR'25})}} & \textbf{56.2} & 50.4 & 66.7 & 39.9 & 50.2 & 59.5 & 39.9 & \textbf{38.0} & 50.1 \\
    GDD \footnotesize{\textit{(R101)}}\normalsize~\cite{gdd}  \textit{\scriptsize{(\textcolor{darkgreen}{CVPR'25})}} & 53.8 & 54.2 & 67.5 & \textbf{45.6} & 52.1 & 60.8 & \textbf{53.9} & 32.4 & 52.5 \\
    \midrule
    \textit{\textbf{Ours\footnotesize{ (Diff. Detector, SD-1.5)}}} & 53.0 & 55.4 & 68.1 & 42.1 & 51.0 & 59.9 & 39.4 & 36.4 & 50.7 \\
    \rowcolor{lightgreen}
    \textit{\textbf{Ours\footnotesize{ (Diff. Guided, R101)}}} & 55.3 & \textbf{62.7} & \textbf{68.2} & 45.5 & \textbf{52.9} & \textbf{62.0} & 48.4 & 37.4 & \textbf{54.1} \\
    \midrule
    \midrule
    \multicolumn{10}{c}{\textit{\textbf{DA methods}}} \\
    MIC~\cite{mic} \textit{\scriptsize{(\textcolor{darkgreen}{CVPR'23})}}& 52.4 & 47.5 & 67.0 & 40.6 & 50.9 & 55.3 & 33.7 & 33.9 & 47.6 \\
    SIGMA++~\cite{li2023sigma++} \textit{\scriptsize{(\textcolor{darkgreen}{TPAMI'23})}}  & 52.2 & 39.9 & 61.0 & 34.8 & 46.4 & 45.1 & 44.6 & 32.1 & 44.5 \\
    CIGAR~\cite{CIGAR} \textit{\scriptsize{(\textcolor{darkgreen}{CVPR'23})}} & 56.6 & 41.3 & 62.1 & 33.7 & 46.1 & 47.3 & 44.3 & 27.8 & 44.9 \\
    CMT~\cite{cao2023cmt} \textit{\scriptsize{(\textcolor{darkgreen}{CVPR'23})}} & \textbf{66.0} & 51.2 & 63.7 & 41.4 & 45.9 & 55.7 & 38.8 & 39.6 & 50.3 \\
    HT~\cite{deng2023HT} \textit{\scriptsize{(\textcolor{darkgreen}{CVPR'23})}} & 55.9 & 50.3 & 67.5 & 40.1 & 52.1 & 55.8 & 49.1 & 32.7 & 50.4 \\
    DSCA~\cite{dsca} \textit{\scriptsize{(\textcolor{darkgreen}{PR'24})}} & 54.7 & 54.9 & 68.4 & 40.1 & 55.4 & 61.0 & 35.9 & 35.1 & 50.7 \\
    UMGA~\cite{umga} \textit{\scriptsize{(\textcolor{darkgreen}{TPAMI'24})}} & 58.0 & 43.7 & 64.9 & 38.3 & 47.9 & 50.1 & 45.6 & 34.8 & 47.9 \\
    CAT~\cite{cat} \textit{\scriptsize{(\textcolor{darkgreen}{CVPR'24})}} & \textbf{66.0} & 53.0 & 63.7 & 44.9 & 44.6 & 57.1 & 49.7 & 40.8 & 52.5 \\
    MTM~\cite{mtm} \textit{\scriptsize{(\textcolor{darkgreen}{AAAI'24})}} & 54.4 & 47.7 & 67.2 & 38.4 & 51.0 & 53.4 & 41.6 & 37.2 & 48.9 \\
    DDT \footnotesize{\textit{(R101)}}\normalsize~\cite{ddt} \textit{\scriptsize{(\textcolor{darkgreen}{MM'24})}} & 53.5 & 52.2 & 64.2 & 43.5 & 50.9 & 60.0 & 42.4 & 33.6 & 50.0 \\
    \midrule
    \rowcolor{lightgreen}
    \textit{\textbf{Ours\footnotesize{ (Diff. Guided, R101)}}} & 56.9 & \textbf{66.2} & \textbf{71.7} & \textbf{49.1} & \textbf{55.6} & \textbf{63.1} & 48.9 & \textbf{41.1} & \textbf{56.6} \\
    \bottomrule
    \end{tabular}%
    }
\end{table}

\begin{table}[ht]
    \centering
    \caption{DG and DA Results (\%) on Clipart, Comic, and Watercolor. }
    \vspace{-4pt}
    \label{tab:voc}
    \begin{minipage}[t]{0.23\textwidth}
        \vspace{0pt}
        \begin{center}
        \renewcommand{\arraystretch}{1.2}
        \setlength{\tabcolsep}{1pt}
        \resizebox{\columnwidth}{!}{%
        \begin{tabular}{l|ccc}
        \toprule
        \multicolumn{4}{c}{\textbf{DG Methods}} \\
        \midrule
        \textbf{Methods} & \textbf{Cli.} & \textbf{Com.} & \textbf{Wat.} \\ \midrule
        Div.~\cite{Diversification} \textit{\scriptsize{(\textcolor{darkgreen}{CVPR'24})}} & 33.7 & 25.5 & 52.5 \\
        DivAlign~\cite{Diversification} \textit{\scriptsize{(\textcolor{darkgreen}{CVPR'24})}} & 38.9 & 33.2 & 57.4 \\
        DDT\footnotesize{ \textit{(SD-1.5)}}\normalsize~\cite{ddt} \textit{\scriptsize{(\textcolor{darkgreen}{MM'24})}} & 47.4 & 44.4 & 58.7 \\
        GDD\footnotesize{ \textit{(SD-1.5)}}\normalsize~\cite{gdd}  \textit{\scriptsize{(\textcolor{darkgreen}{CVPR'25})}} & 58.3 & 51.9 & 68.4 \\
        GDD\footnotesize{ \textit{(R101)}}\normalsize~\cite{gdd}  \textit{\scriptsize{(\textcolor{darkgreen}{CVPR'25})}} & 40.8 & 29.7 & 54.2 \\
        \midrule
        \rowcolor{lightgreen}
        \textit{\textbf{Ours\footnotesize{ (Diff. Detector, SD-1.5)}}} & \textbf{64.1} & \textbf{55.2} & \textbf{69.7} \\
        
        \textit{\textbf{Ours\footnotesize{ (Diff. Guided, R101)}}} & 40.5 & 30.0 & 56.6 \\ \bottomrule
        \end{tabular}%
        }
        \end{center}
    \end{minipage}%
    \hfill
    \begin{minipage}[t]{0.23\textwidth}
        \vspace{0pt}
        \begin{center}
        \setlength{\tabcolsep}{2pt}
        \resizebox{\columnwidth}{!}{%
        \begin{tabular}{l|ccc}
        \toprule
        \multicolumn{4}{c}{\textbf{DA Methods}} \\
        \midrule
        \textbf{Methods} & \textbf{Cli.} & \textbf{Com.} & \textbf{Wat.} \\ \midrule
        SWDA~\cite{saito2019swda} \textit{\scriptsize{(\textcolor{darkgreen}{CVPR'19})}} & 38.1 & 29.4 & 53.3 \\
        UMT~\cite{umt} \textit{\scriptsize{(\textcolor{darkgreen}{CVPR'21})}} & 44.1 & -- & 58.1 \\
        SADA~\cite{sada} \textit{\scriptsize{(\textcolor{darkgreen}{IJCV'21})}} & 43.3 & -- & 56.0 \\
        DBGL~\cite{dbgl} \textit{\scriptsize{(\textcolor{darkgreen}{ICCV'21})}} & 41.6 & 29.7 & 53.8 \\
        AT~\cite{AT} \textit{\scriptsize{(\textcolor{darkgreen}{CVPR'22})}} & 49.3 & -- & 59.9 \\
        D-ADAPT~\cite{dadapt} \textit{\scriptsize{(\textcolor{darkgreen}{ICLR'22})}} & 49.0 & 40.5 & -- \\
        TIA~\cite{tia} \textit{\scriptsize{(\textcolor{darkgreen}{CVPR'22})}} & 46.3 & -- & -- \\
        LODS~\cite{lods} \textit{\scriptsize{(\textcolor{darkgreen}{CVPR'22})}} & 45.2 & -- & 58.2 \\
        CIGAR~\cite{CIGAR} \textit{\scriptsize{(\textcolor{darkgreen}{CVPR'23})}} & 46.2 & -- & -- \\
        CMT~\cite{cao2023cmt} \textit{\scriptsize{(\textcolor{darkgreen}{CVPR'23})}} & 47.0 & -- & -- \\ 
        DAVimNet~\cite{davimnet} \textit{\scriptsize{(\textcolor{darkgreen}{ArXiv'24})}} & 43.8 & -- & 54.8 \\
        UMGA~\cite{umga} \textit{\scriptsize{(\textcolor{darkgreen}{TPAMI'24})}} & 49.9 & -- & 62.1 \\
        CAT~\cite{cat} \textit{\scriptsize{(\textcolor{darkgreen}{CVPR'24})}} & 49.1 & -- & -- \\
        DDT \footnotesize{\textit{(R101)}}\normalsize~\cite{ddt} \textit{\scriptsize{(\textcolor{darkgreen}{MM'24})}} & 55.6 & 50.2 & 63.7 \\
        \midrule
        \rowcolor{lightgreen}
        \textit{\textbf{Ours\footnotesize{ (Diff. Guided, R101)}}} & \textbf{58.2} & \textbf{50.5} & \textbf{68.0} \\ \bottomrule
        \end{tabular}%
        }
        \end{center}
    \end{minipage}
\end{table}

\begin{table}[ht]
    \centering
    \caption{DG Results (\%) on Diverse Weather Datasets.}
    \vspace{-4pt}
    \label{tab:dwd}
    \setlength{\tabcolsep}{8pt}
    \resizebox{1\columnwidth}{!}{%
    \begin{tabular}{l|cccc|>{\columncolor{gray!20}}c}
    \toprule
    \textbf{Methods} & \textbf{DF} & \textbf{DR} & \textbf{NR} & \textbf{NS} & \textbf{Average} \\ \midrule
    CDSD~\cite{cdsd} \textit{\scriptsize{(\textcolor{darkgreen}{CVPR'22})}} & 33.5 & 28.2 & 16.6 & 36.6 & 28.7 \\
    SHADE~\cite{shade} \textit{\scriptsize{(\textcolor{darkgreen}{ECCV'22})}} & 33.4 & 29.5 & 16.8 & 33.9 & 28.4 \\
    CLIPGap~\cite{clip_gap} \textit{\scriptsize{(\textcolor{darkgreen}{CVPR'23})}} & 32.0 & 26.0 & 12.4 & 34.4 & 26.2 \\
    SRCD~\cite{srcd} \textit{\scriptsize{(\textcolor{darkgreen}{TNNLS'24})}} & 35.9 & 28.8 & 17.0 & 36.7 & 29.6 \\
    G-NAS~\cite{gnas} \textit{\scriptsize{(\textcolor{darkgreen}{AAAI'24})}} & 36.4 & 35.1 & 17.4 & 45.0 & 33.5 \\
    PhysAug~\cite{physaug} \textit{\scriptsize{(\textcolor{darkgreen}{ArXiv'24})}} & 40.8 & 41.2 & 23.1 & 44.9 & 37.5 \\ 
    OA-DG~\cite{oamix} \textit{\scriptsize{(\textcolor{darkgreen}{AAAI'24})}} & 38.3 & 33.9 & 16.8 & 38.0 & 31.8 \\
    DivAlign~\cite{Diversification} \textit{\scriptsize{(\textcolor{darkgreen}{CVPR'24})}} & 37.2 & 38.1 & 24.1 & 42.5 & 35.5 \\
    UFR~\cite{ufr} \textit{\scriptsize{(\textcolor{darkgreen}{CVPR'24})}} & 39.6 & 33.2 & 19.2 & 40.8 & 33.2 \\
    Prompt-D~\cite{prompt-d} \textit{\scriptsize{(\textcolor{darkgreen}{CVPR'24})}} & 39.1 & 33.7 & 19.2 & 38.5 & 32.6 \\
    DIDM~\cite{didm} \textit{\scriptsize{(\textcolor{darkgreen}{ArXiv'25})}} & 39.3 & 35.4 & 19.2 & 42.0 & 34.0 \\
    GDD \footnotesize{\textit{(SD-1.5)}}\normalsize~\cite{gdd}  \textit{\scriptsize{(\textcolor{darkgreen}{CVPR'25})}} & 43.3 & 42.5 & 27.8 & 47.0 & 40.2 \\
    GDD \footnotesize{\textit{(R101)}}\normalsize~\cite{gdd}  \textit{\scriptsize{(\textcolor{darkgreen}{CVPR'25})}} & 44.7 & 37.4 & 21.7 & 48.7 & 38.1 \\
    \midrule
    \rowcolor{lightgreen}
    \textit{\textbf{Ours\footnotesize{ (Diff. Detector, SD-1.5)}}} & \textbf{48.5} & \textbf{48.4} & \textbf{31.3} & \textbf{51.6} & \textbf{45.0} \\
    \textit{\textbf{Ours\footnotesize{ (Diff. Guided, R101)}}} & 46.7 & 39.1 & 22.4 & 50.5 & 39.7 \\ \bottomrule
    \end{tabular}%
    }
\end{table}

 \begin{table*}[ht]
    \centering
    \caption{Generalization Detection Results (\%) on Cityscapes-Corruption Benchmark.}
    \vspace{-4pt}
    \label{tab:cityscapes-c}
    \setlength{\tabcolsep}{4pt}
    \renewcommand{\arraystretch}{1.0}
    \resizebox{2.1\columnwidth}{!}{%
    \begin{tabular}{l|c|ccc|cccc|ccc|ccccc|>{\columncolor{gray!20}}c}
    \toprule
    & & \multicolumn{3}{c|}{\textbf{Noise}} & \multicolumn{4}{c|}{\textbf{Blur}} & \multicolumn{3}{c|}{\textbf{Weather}} & \multicolumn{5}{c|}{\textbf{Digital}} & \\
    \textbf{Methods} & \textcolor{gray}{Clean} & Gauss. & Shot & Impulse & Defocus & Glass & Motion & Zoom & Snow & Frost & Fog & Bright & Contrast & Elastic & JPEG & Pixel & \textbf{mPC \textcolor{red}{$\uparrow$}} \\
    \midrule
    FSCE~\cite{fsce} \textit{\scriptsize{(\textcolor{darkgreen}{CVPR'21})}} & \textcolor{gray}{43.1} & 7.4 & 10.2 & 8.2 & 23.3 & 20.3 & 21.5 & 4.8 & 5.6 & 23.6 & 37.1 & 38.0 & 31.9 & \textbf{40.2} & 20.4 & 23.2 & 21.0 \\
    OA-Mix~\cite{oamix} \textit{\scriptsize{(\textcolor{darkgreen}{AAAI'24})}} & \textcolor{gray}{42.7} & 7.2 & 9.6 & 7.7 & 22.8 & 18.8 & 21.9 & \textbf{5.4} & 5.2 & 23.6 & 37.3 & 38.7 & 31.9 & \textbf{40.2} & 20.2 & 22.2 & 20.8 \\
    OA-DG~\cite{oamix} \textit{\scriptsize{(\textcolor{darkgreen}{AAAI'24})}} & \textcolor{gray}{\textbf{43.4}} & 8.2 & 10.6 & 8.4 & 24.6 & 20.5 & 22.3 & 4.8 & 6.1 & 25.0 & 38.4 & 39.7 & 32.8 & \textbf{40.2} & 22.0 & 23.8 & 21.8 \\
    GDD \footnotesize{\textit{(SD-1.5)}}\normalsize~\cite{gdd}  \textit{\scriptsize{(\textcolor{darkgreen}{CVPR'25})}} & \textcolor{gray}{34.7} & 20.3 & 23.2 & 17.2 & 26.8 & 21.7 & 23.7 & 3.4 & 16.6 & 24.2 & 32.5 & 34.4 & 30.6 & 33.7 & 29.1 & 24.4 & 24.1 \\
    GDD \footnotesize{\textit{(R50)}}\normalsize~\cite{gdd}  \textit{\scriptsize{(\textcolor{darkgreen}{CVPR'25})}} & \textcolor{gray}{42.1} & 11.0 & 13.6 & 10.8 & 25.0 & 14.2 & 21.4 & 3.4 & 5.4 & 24.0 & 39.6 & 40.3 & \textbf{36.3} & 39.2 & 18.9 & 16.0 & 21.3 \\
    \midrule
    \rowcolor{lightgreen}
    \textit{\textbf{Ours\footnotesize{ (Diff. Detector, SD-1.5)}}} & \textcolor{gray!50}{40.1} & \textbf{23.1} & \textbf{26.6} & \textbf{20.6} & \textbf{29.7} & \textbf{24.5} & \textbf{25.5} & 4.1 & \textbf{18.3} & \textbf{28.2} & 37.2 & 39.2 & 35.5 & 37.4 & \textbf{32.7} & \textbf{28.8} & \textbf{27.4} \\ 
    \textit{\textbf{Ours\footnotesize{ (Diff. Guided, R50)}}} & \textcolor{gray}{41.5} & 13.9 & 16.6 & 13.5 & 24.8 & 14.2 & 22.0 & 3.7 & 5.9 & 24.7 & \textbf{40.0} & \textbf{40.4} & 35.9 & 39.5 & 18.9 & 17.1 & 22.1 \\
    \bottomrule
    \end{tabular}%
    }
\end{table*}

\begin{table*}[h]
    \centering
    \caption{Comparison of DG Performance on COCO Generalization Benchmark Under Different Training Data Scales. \small{\textit{* indicates we don't apply object alignment in consistency loss in FCOS and DINO detectors.}}}
    \vspace{-4pt}
    \label{tab:coco}
    \renewcommand{\arraystretch}{1.0}
    \resizebox{2.1\columnwidth}{!}{%
    \setlength{\tabcolsep}{6pt}
    \begin{tabular}{l|l|c|ccc|c|ccc|cccc|>{\columncolor{gray!20}}c|>{\columncolor{gray!10}}c}
    \toprule
    \textbf{Settings} & \textbf{Models} & \textbf{\textcolor{gray}{COCO Val}} & \textbf{VOC} & \textbf{City.} & \textbf{BDD.} & \textbf{Foggy.} & \textbf{Cli.} & \textbf{Com.} & \textbf{Wat.} & \textbf{DF} & \textbf{DR} & \textbf{NR} & \textbf{NS} & \textbf{Avg. on 11} & \textbf{Inf. time (ms)} \\ 
    \midrule
    \multirow{7}{*}{\shortstack{\textbf{{1\%}} COCO,\\Faster R-CNN}} 
    & ResNet\footnotesize{-R50}\normalsize~\cite{resnet} & \textcolor{gray}{20.4} & 36.4 & 26.6 & 18.6 & 8.2 & 10.1 & 6.6 & 14.3 & 10.6 & 5.9 & 1.1 & 6.5 & 13.2 & - \\
    & ConvNeXt\footnotesize{-Base}\normalsize~\cite{Convnext} & \textcolor{gray}{32.5} & 53.9 & 34.9 & \textbf{28.2} & 29.8 & 18.8 & 14.9 & 26.9 & 25.6 & 16.2 & 5.9 & 18.6 & 24.9 & - \\
    & Swin\footnotesize{-Base}\normalsize~\cite{swin} & \textcolor{gray}{27.7} & 46.0 & 29.0 & 23.1 & 16.3 & 12.9 & 8.1 & 19.8 & 16.8 & 10.3 & 2.1 & 9.0 & 17.6 & - \\
    & VIT\footnotesize{-Base}\normalsize~\cite{vit} & \textcolor{gray}{28.1} & 53.3 & 18.0 & 18.0 & 8.8 & 11.1 & 6.1 & 11.4 & 13.8 & 11.1 & 2.6 & 6.9 & 14.6 & - \\
    & GLIP \footnotesize{(Swin-Tiny)}\normalsize~\cite{glip} & \textcolor{gray}{32.5} & 53.9 & 36.1 & 27.5 & 19.6 & 18.6 & 13.0 & 16.1 & 20.1 & 14.5 & 4.1 & 13.7 & 21.6 & - \\
    \rowcolor{lightgreen}
    & \textit{\textbf{Ours\footnotesize{ \textit{(SD-1.5)}}}} & \textcolor{gray}{\textbf{33.4}} & \textbf{70.4} & \textbf{36.5} & 27.9 & \textbf{31.0} & \textbf{44.0} & \textbf{37.8} & \textbf{50.1} & \textbf{26.6} & \textbf{19.8} & \textbf{10.7} & \textbf{20.8} & \textbf{34.1} & - \\ 
    \midrule
    \midrule
    \multirow{7}{*}{\shortstack{{\textbf{{100\%}}} COCO,\\Faster R-CNN}} 
    & ResNet\footnotesize{-R50}\normalsize~\cite{resnet} & \textcolor{gray}{58.1} & 84.0 & 49.5 & 45.8 & 35.2 & 32.6 & 23.7 & 40.8 & 33.3 & 24.8 & 10.7 & 30.3 & 37.3 & 27 \\
    & ConvNeXt\footnotesize{-Base}\normalsize~\cite{Convnext} & \textcolor{gray}{64.5} & 83.4 & 53.4 & 50.9 & 43.1 & 43.5 & 35.5 & 47.9 & 40.5 & 34.4 & 17.6 & 36.0 & 44.2 & 54 \\
    & Swin\footnotesize{-Base}\normalsize~\cite{swin} & \textcolor{gray}{61.5} & 79.6 & 52.9 & 48.0 & 38.9 & 37.8 & 29.4 & 41.7 & 37.6 & 32.1 & 14.4 & 33.0 & 40.5 & 55 \\
    & VIT\footnotesize{-Base}\normalsize~\cite{vit} & \textcolor{gray}{62.7} & 86.3 & 35.1 & 40.4 & 22.9 & 38.3 & 24.1 & 48.9 & 31.2 & 29.1 & 12.4 & 26.3 & 35.9 & 78 \\
    & GLIP \footnotesize{(Swin-Tiny)}\normalsize~\cite{glip} & \textcolor{gray}{62.0} & 79.9 & 54.0 & 48.6 & 41.4 & 38.9 & 29.6 & 40.3 & 37.9 & 31.5 & 15.0 & 34.9 & 41.1 & 31 \\
    \rowcolor{lightgreen}
    & \textit{\textbf{Ours\footnotesize{ \textit{(SD-1.5)}}}} & \textcolor{gray}{\textbf{67.0}} & \textbf{86.6} & \textbf{54.1} & \textbf{51.2} & \textbf{45.8} & \textbf{64.6} & \textbf{51.9} & \textbf{66.6} & \textbf{41.7} & \textbf{39.2} & \textbf{21.7} & \textbf{40.7} & \textbf{51.3} & 164 \\ 
    \midrule
    \multirow{2}{*}{\shortstack{{\textbf{{100\%}}} COCO,\\\textit{SD-1.5}}} 
    & FCOS*~\cite{fcos} & \textcolor{gray}{64.3} & 85.0 & 50.3 & 48.3 & 42.0 & 62.9 & 52.1 & 65.9 & 39.6 & 36.4 & 19.7 & 39.0 & 49.2 & 171 \\
    & DINO*~\cite{dino_detector} & \textcolor{gray}{68.3} & 88.2 & 51.2 & 51.1 & 45.9 & 61.7 & 50.8 & 63.6 & 42.0 & 39.3 & 22.0 & 40.2 & 50.5 & 232 \\
    \bottomrule
    \end{tabular}%
    }
\end{table*}

We present our DA results in Tab.~\ref{tab:bdd100k},~\ref{tab:FoggyCityscapes}, and~\ref{tab:voc}, while DG results are shown in Tab.~\ref{tab:bdd100k},~\ref{tab:FoggyCityscapes},~\ref{tab:voc},~\ref{tab:dwd}, and~\ref{tab:cityscapes-c}, with comparisons to SOTA methods. The \textbf{bold} values indicate the best results, and \colorbox{lightgreen}{\small{Yellow Background}} highlights methods achieving the best average performance.
To save space in tables, we use the following dataset abbreviations: \textbf{City.}~(Cityscapes), \textbf{BDD.}~(BDD100K), \textbf{Foggy.}~(FoggyCityscapes), \textbf{Cli.}~(Clipart), \textbf{Com.}~(Comic), \textbf{Wat.}~(Watercolor), \textbf{DF}~(Daytime-Foggy), \textbf{DR}~(Dusk-Rainy), \textbf{NR}~(Night-Rainy), and \textbf{NS}~(Night-Sunny).

\subsubsection{DG Results of Diff. Detector} 

Tab.~\ref{tab:bdd100k},~\ref{tab:FoggyCityscapes},~\ref{tab:voc},~\ref{tab:dwd}, and~\ref{tab:cityscapes-c} show diff. detector outperforms SOTA GDD~\cite{gdd} across all benchmarks. Both approaches significantly surpass other recent methods~\cite{mad,srcd,Diversification,clip_gap,gnas,oamix,ufr,prompt-d}, confirming diffusion models' effectiveness for DG. Notably, our method achieves \textbf{3.0\%} average mAP improvement while reducing inference time by \textbf{75\%} compared to GDD~\cite{gdd} as shown in Fig.~\ref{fig:comparisons}.

\subsubsection{DG Results of Diff. Guided Detector} 
Tab.~\ref{tab:bdd100k},~\ref{tab:FoggyCityscapes},~\ref{tab:voc},~\ref{tab:dwd}, and~\ref{tab:cityscapes-c} show the results of ordinary detectors (Faster R-CNN with R101) guided by diff. detector. Following the settings in GDD~\cite{gdd}, our methods achieve improvements \textbf{\{0.4, 1.6, 0.8, 1.8, 1.8\}}\% mAP improvements on 5 DG benchmarks compared to GDD, indicating that a stronger diff. detector typically provides better guidance.

\subsubsection{DA Results of Diff. Guided Detector}
Similarly, we adopt the diff. detector as a teacher model to generate pseudo-labels on unlabeled target domains and guides the student model through semi-supervised learning. Following the same settings as DDT~\cite{ddt}, our method achieves \textbf{\{7.9, 6.6, 1.7\}}\% improvements on 3 DA benchmarks.

\subsubsection{COCO Benchmark Evaluation}

As shown in Tab.~\ref{tab:coco}, our diff. detector outperforms~\cite{resnet,Convnext,swin,vit,glip} on the \textbf{COCO Generalization Benchmark} using both 1\% and 100\% training data. Our method excels particularly in \textbf{Data-Scarce Scenarios}, \textbf{Extreme Domain Shifts} (Clipart, Comic, Watercolor~\cite{clipart_comic_watercolor}, and Night-Rainy), demonstrating its suitability for limited data and substantial domain gaps. Results remain consistent across different detectors (\textbf{FCOS}~\cite{fcos} and \textbf{DINO}~\cite{dino_detector}), confirming our framework's broad applicability.

\subsection{{Failure Cases Analysis}}

Although diff. detector and diff. guided detector achieved significant improvements on DG and DA benchmarks, the ordinary detector guided by diff. detector showed limited gains in \textit{Real to Artistic} scenarios, underperforming compared to GDD~\cite{gdd} (by \textcolor{red}{-0.3\%} on Clipart) and DivAlign~\cite{Diversification} (by \textcolor{red}{-3.2\%} on Comic and \textcolor{red}{-0.8\%} on Watercolor). While generating pseudo-labels works efficiently for DA tasks, frameworks that align solely on source domains may fail when target domains remain unseen with enormous domain gaps.

\section{Ablation Studies}

\begin{table*}[!h]
    \begin{minipage}[t]{0.7\textwidth}
    \centering
    \caption{Ablation Studies of Proposed Components on \textbf{\textit{Diff. Detector}}. \textit{\small{* indicates we test the inference time by applying same settings on 4090 GPU with scale (1333, 800), which may different from GDD~\cite{gdd} reported.}}}
    \vspace{-4pt}
    \label{tab:diff-detector}
    \setlength{\tabcolsep}{3pt}
    \resizebox{1.0\columnwidth}{!}{%
    \begin{tabular}{c|c|cc|cc|cc|ccc|c}
    \toprule
    \multirow{2}{*}{\textbf{Detector}} & 
    \multirow{2}{*}{\begin{tabular}[c]{@{}c@{}}\small\textbf{Domain}\\\small\textbf{Aug.}\end{tabular}} & 
    \multirow{2}{*}{\begin{tabular}[c]{@{}c@{}}\small\textbf{Feature}\\\small\textbf{Coll.}\end{tabular}} & 
    \multirow{2}{*}{\begin{tabular}[c]{@{}c@{}}\small\textbf{Feature}\\\small\textbf{Fusion}\end{tabular}} & 
    \multirow{2}{*}{\begin{tabular}[c]{@{}c@{}}\small\textbf{Aux.}\\\small\textbf{Branch}\end{tabular}} & 
    \multirow{2}{*}{\begin{tabular}[c]{@{}c@{}}\small\textbf{Consist.}\\\small\textbf{Loss}\end{tabular}} &
    \multicolumn{2}{c|}{\textbf{In-Domain}} & \multicolumn{3}{c|}{\textbf{Cross-Domain}} & \multirow{2}{*}{\begin{tabular}[c]{@{}c@{}}\textbf{Inf. time}\\\textbf{(ms)}\end{tabular}} \\
     & & & & & & City. & VOC & Foggy. & BDD. & Cli. & \\
    \midrule
    \textbf{w/o} \textit{Noise Adding} & - & - & - & - & - & 37.4 & 74.9 & 30.8 & 29.1 & 38.2 & \cellcolor{gray!20}132 \\
    \textbf{w/}~~\textit{Noise Adding} & - & - & - & - & - & 38.5 & 74.7 & 34.9 & 34.3 & 43.6 & \cellcolor{gray!20}132 \\
    \midrule
    \multirow{6}{*}{\begin{tabular}[c]{@{}c@{}}\textbf{Ours ($T=1$)}\\\textbf{w/}~~\textit{Noise Adding}\end{tabular}}
    & \checkmark & - & - & - & - & 40.9\footnotesize{\textcolor{red}{+2.4}} & 75.2\footnotesize{\textcolor{red}{+0.5}} & 39.4\footnotesize{\textcolor{red}{+4.5}} & 36.1\footnotesize{\textcolor{red}{+1.8}} & 47.2\footnotesize{\textcolor{red}{+3.6}} & \cellcolor{gray!20}132 \\
    & - & \checkmark & - & - & - & 44.3\footnotesize{\textcolor{red}{+5.8}} & 77.3\footnotesize{\textcolor{red}{+2.6}} & 37.8\footnotesize{\textcolor{red}{+2.9}} & 35.2\footnotesize{\textcolor{red}{+0.9}} & 46.6\footnotesize{\textcolor{red}{+3.0}} & \cellcolor{gray!20}144 \\
    & - & - & \checkmark & - & - & 46.1\footnotesize{\textcolor{red}{+7.6}} & 79.4\footnotesize{\textcolor{red}{+4.7}} & 38.6\footnotesize{\textcolor{red}{+3.7}} & 36.1\footnotesize{\textcolor{red}{+1.8}} & 49.4\footnotesize{\textcolor{red}{+5.8}} & \cellcolor{gray!20}151 \\
    & - & \checkmark & \checkmark & - & - & 52.4\footnotesize{\textcolor{red}{+13.9}} & 81.3\footnotesize{\textcolor{red}{+6.6}} & 40.1\footnotesize{\textcolor{red}{+5.2}} & 37.9\footnotesize{\textcolor{red}{+3.6}} & 51.2\footnotesize{\textcolor{red}{+7.6}} & \cellcolor{gray!20}164 \\
    \cmidrule{2-12}
    & - & - & - & \checkmark & - & 46.9\footnotesize{\textcolor{red}{+8.4}} & 78.9\footnotesize{\textcolor{red}{+4.2}} & 40.4\footnotesize{\textcolor{red}{+5.5}} & 39.0\footnotesize{\textcolor{red}{+4.7}} & 53.3\footnotesize{\textcolor{red}{+9.7}} & \cellcolor{gray!20}164 \\
    & - & - & - & \checkmark & \checkmark & 46.0\footnotesize{\textcolor{red}{+7.5}} & 78.8\footnotesize{\textcolor{red}{+4.1}} & 43.2\footnotesize{\textcolor{red}{+8.3}} & 41.3\footnotesize{\textcolor{red}{+7.0}} & 57.6\footnotesize{\textcolor{red}{+14.0}} & \cellcolor{gray!20}164 \\
    \cmidrule{2-12}
    & \checkmark & \checkmark & \checkmark & \checkmark & - & \textbf{62.4} & \textbf{87.9} & 45.8 & 44.7 & 59.8 & \cellcolor{gray!20}164 \\
    & \checkmark & \checkmark & \checkmark & \checkmark & \checkmark & 61.1\footnotesize{\textcolor{darkgreen}{-1.3}} & 87.3\footnotesize{\textcolor{darkgreen}{-0.6}} & \textbf{50.7}\footnotesize{\textcolor{red}{+4.9}} & \textbf{49.3}\footnotesize{\textcolor{red}{+4.6}} & \textbf{64.1}\footnotesize{\textcolor{red}{+4.3}} & \cellcolor{gray!20}164 \\
    \midrule
    \multirow{2}{*}{\textbf{GDD}~\cite{gdd}} 
    & \multicolumn{5}{c|}{$T=5$,  \textbf{w/}~~\textit{Noise Adding}} & 59.8 & 84.8 & 50.1 & 46.6 & 58.3 & \footnotesize \cellcolor{gray!20}789 / \textbf{679}* \\
    & \multicolumn{5}{c|}{$T=1$,  \textbf{w/o} \textit{Noise Adding}} & 36.8\footnotesize{\textcolor{darkgreen}{-23.0}} & 74.2\footnotesize{\textcolor{darkgreen}{-10.6}} & 30.8\footnotesize{\textcolor{darkgreen}{-19.3}} & 28.6\footnotesize{\textcolor{darkgreen}{-18.0}} & 37.4\footnotesize{\textcolor{darkgreen}{-20.9}} & \footnotesize \cellcolor{gray!20}270 / \textbf{194}* \\
    \bottomrule
    \end{tabular}%
    }    
    \end{minipage}
    \hfill
    \begin{minipage}[t]{0.26\textwidth}
    \centering
    \caption{Studies of loss weights $\gamma$ in $\mathcal{L}_{cons}$ (Equ.~\ref{l_cons})  and $\lambda$ in $\mathcal{L}_{total}$ (Equ.~\ref{l_total}).}
    \vspace{-4pt}
    \label{tab:loss_weights}
    \setlength{\tabcolsep}{5pt}
    \renewcommand{\arraystretch}{1.18}
    \resizebox{0.66\columnwidth}{!}{%
    \begin{tabular}{c|ccc}
        \toprule
        $\gamma$ & \textbf{Foggy.} & \textbf{BDD.} & \textbf{Cli.} \\
        \midrule
        0.0 & 47.6 & 45.9 & 61.2 \\
        0.5 & 50.1 & 49.0 & 63.7 \\
        1.0 & 50.7 & \textbf{49.3} & 64.1 \\
        1.5 & \textbf{50.8} & \textbf{49.3} & \textbf{64.2} \\
        2.0 & 49.1 & 49.1 & 62.7 \\
        \midrule
        \midrule
        $\lambda$ & \textbf{Foggy.} & \textbf{BDD.} & \textbf{Cli.} \\
        \midrule
        0.0 & 45.8 & 44.7 & 59.8 \\
        0.5 & 50.3 & 48.9 & 63.1 \\
        1.0 & \textbf{50.7} & 49.3 & \textbf{64.1} \\
        1.5 & 50.6 & \textbf{49.6} & 64.0 \\
        2.0 & 50.1 & 48.1 & 63.8 \\
        \bottomrule
    \end{tabular}%
    }
    \end{minipage}
\end{table*}

\begin{table}[!ht]
    \centering
    \caption{Comprehensive Results (\%) of \textit{\textbf{SD-2.1}} Version.}
    \vspace{-4pt}
    \label{tab:sd-version}
    \setlength{\tabcolsep}{4pt}
    \resizebox{\columnwidth}{!}{%
    \begin{tabular}{l|ccccccccc}
    \toprule
    \textbf{Methods} & \textbf{BDD.} & \textbf{Foggy.} & \textbf{Cli.} & \textbf{Com.} & \textbf{Wat.} & \textbf{DF} & \textbf{DR} & \textbf{NR} & \textbf{NS} \\
    \midrule
    \multicolumn{10}{c}{\textit{Diff. Detector, DG Settings}} \\
    DDT \footnotesize{\textit{(SD-2.1)}}~\cite{ddt}  & 34.6 & - & 45.4 & 42.8 & 58.7 & - & - & - & - \\
    GDD \footnotesize{\textit{(SD-2.1)}}~\cite{gdd}  & 45.8 & 48.3 & 51.7 & 46.6 & 62.1 & 44.6 & 41.6 & 23.2 & 46.4 \\
    \textit{\textbf{Ours \footnotesize{\textit{(SD-2.1)}}}} & \textbf{48.0} & \textbf{50.3} & \textbf{59.7} & \textbf{54.5} & \textbf{68.6} & \textbf{48.7} & \textbf{47.3} & \textbf{29.8} & \textbf{51.8} \\
    \footnotesize{\textcolor{red}{+Gain}} & \footnotesize{\textcolor{red}{+2.2}} & \footnotesize{\textcolor{red}{+2.0}} & \footnotesize{\textcolor{red}{+8.0}} & \footnotesize{\textcolor{red}{+7.9}} & \footnotesize{\textcolor{red}{+6.5}} & \footnotesize{\textcolor{red}{+4.1}} & \footnotesize{\textcolor{red}{+5.7}} & \footnotesize{\textcolor{red}{+6.6}} & \footnotesize{\textcolor{red}{+5.4}} \\
    \midrule
    \multicolumn{10}{c}{\textit{Diff. Guided Detector, DG Settings}} \\
    GDD \footnotesize{\textit{(SD-2.1)}}\normalsize~\cite{gdd} & 46.1 & 51.0 & 32.7 & 24.9 & 50.6 & 44.7 & 37.1 & 20.0 & 49.3 \\
    \textit{\textbf{Ours \footnotesize{\textit{(SD-2.1)}}}} & \textbf{46.5} & \textbf{54.4} & \textbf{38.4} & \textbf{30.0} & \textbf{56.3} & \textbf{46.7} & \textbf{39.1} & \textbf{22.4} & \textbf{50.5} \\
    \footnotesize{\textcolor{red}{+Gain}} & \footnotesize{\textcolor{red}{+0.4}} & \footnotesize{\textcolor{red}{+3.4}} & \footnotesize{\textcolor{red}{+5.7}} & \footnotesize{\textcolor{red}{+5.1}} & \footnotesize{\textcolor{red}{+5.7}} & \footnotesize{\textcolor{red}{+2.0}} & \footnotesize{\textcolor{red}{+2.0}} & \footnotesize{\textcolor{red}{+2.4}} & \footnotesize{\textcolor{red}{+1.2}} \\
    \midrule
    \multicolumn{10}{c}{\textit{Diff. Guided Detector, DA Settings}} \\
    DDT \footnotesize{\textit{(SD-2.1)}}~\cite{ddt}  & 42.3 & - & 53.7 & 48.9 & 63.3 & - & - & - & - \\
    \textit{\textbf{Ours \footnotesize{\textit{(SD-2.1)}}}} & \textbf{51.7} & \textbf{56.6} & \textbf{56.3} & \textbf{50.2} & \textbf{66.2} & - & - & - & - \\
    \footnotesize{\textcolor{red}{+Gain}} & \footnotesize{\textcolor{red}{+9.4}} & - & \footnotesize{\textcolor{red}{+2.6}} & \footnotesize{\textcolor{red}{+1.3}} & \footnotesize{\textcolor{red}{+2.9}} & - & - & - & - \\
    \bottomrule
    \end{tabular}%
    }
    \end{table}

\begin{table}[!ht]
    \begin{minipage}{\columnwidth}
        \centering  
        \caption{Ablation Study on Diff. Guided Detector for \textbf{DG}.}
        \vspace{-4pt}
        \label{tab:DG}
        \setlength{\tabcolsep}{1pt}
        \resizebox{\columnwidth}{!}{%
        \begin{tabular}{ccc|cccccc}
        \toprule
        \multirow{3}{*}{\footnotesize\begin{tabular}[c]{@{}c@{}}\textbf{Domain}\\\textbf{Aug.}\end{tabular}} & 
        \multirow{3}{*}{\footnotesize\begin{tabular}[c]{@{}c@{}}\textbf{Feature}\\\textbf{Alignment}\end{tabular}} & 
        \multirow{3}{*}{\footnotesize\begin{tabular}[c]{@{}c@{}}\textbf{Object}\\\textbf{Alignment}\end{tabular}} &
        \multicolumn{6}{c}{\textbf{Unseen Target Domain}} \\
        \cmidrule(lr){4-9}
         &  &  & \textbf{BDD.} & \textbf{Foggy.} & \textbf{DF} & \textbf{DR} & \textbf{NR} & \textbf{NS} \\
        \midrule
        - & - & - & 25.4 & 30.7 & 28.8 & 24.1 & 12.4 & 31.4 \\
        \checkmark & - & - & 36.2\footnotesize{\textcolor{red}{+10.8}} & 47.9\footnotesize{\textcolor{red}{+17.2}} & 39.9\footnotesize{\textcolor{red}{+11.1}} & 34.8\footnotesize{\textcolor{red}{+10.7}} & 16.1\footnotesize{\textcolor{red}{+3.7}} & 41.2\footnotesize{\textcolor{red}{+9.8}} \\
        \checkmark & \checkmark & - & 39.5\footnotesize{\textcolor{red}{+3.3}} & 48.9\footnotesize{\textcolor{red}{+1.0}} & 42.1\footnotesize{\textcolor{red}{+2.2}} & 35.2\footnotesize{\textcolor{red}{+0.4}} & 18.2\footnotesize{\textcolor{red}{+2.1}} & 43.0\footnotesize{\textcolor{red}{+1.8}} \\
        \checkmark & \checkmark & \checkmark & \textbf{46.7}\footnotesize{\textcolor{red}{+7.2}} & \textbf{54.1}\footnotesize{\textcolor{red}{+5.2}} & \textbf{46.7}\footnotesize{\textcolor{red}{+4.6}} & \textbf{39.1}\footnotesize{\textcolor{red}{+3.9}} & \textbf{22.4}\footnotesize{\textcolor{red}{+4.2}} & \textbf{50.5}\footnotesize{\textcolor{red}{+7.5}} \\
        \bottomrule
        \end{tabular}%
        }
    \end{minipage}
    
    \vspace{15pt}
    
    \begin{minipage}{\columnwidth}
        \centering
        \caption{Ablation Study on Diff. Guided Detector for \textbf{DA}.}
        \vspace{-4pt}
        \label{tab:DA}
        \setlength{\tabcolsep}{3pt}
        \resizebox{\columnwidth}{!}{%
        \begin{tabular}{ccc|ccccc}
        \toprule
        \multirow{3}{*}{\footnotesize\begin{tabular}[c]{@{}c@{}}\textbf{Domain}\\\textbf{Aug.}\end{tabular}} & 
        \multirow{3}{*}{\footnotesize\begin{tabular}[c]{@{}c@{}}\textbf{Feature}\\\textbf{Alignment}\end{tabular}} & 
        \multirow{3}{*}{\footnotesize\begin{tabular}[c]{@{}c@{}}\textbf{Object}\\\textbf{Alignment}\end{tabular}} &
        \multicolumn{5}{c}{\textbf{Unlabeled Target Domain}} \\
        \cmidrule(lr){4-8}
         &  &  & \textbf{BDD.} & \textbf{Foggy.} & \textbf{Cli.} & \textbf{Com.} & \textbf{Wat.} \\
        \midrule
        - & - & - & 25.4 & 30.7 & 27.2 & 18.1 & 41.5 \\
        \checkmark & - & - & 38.1\footnotesize{\textcolor{red}{+12.7}} & 48.9\footnotesize{\textcolor{red}{+18.2}} & 39.4\footnotesize{\textcolor{red}{+12.2}} & 27.4\footnotesize{\textcolor{red}{+9.3}} & 50.9\footnotesize{\textcolor{red}{+9.4}} \\
        \checkmark & \checkmark & - & 40.2\footnotesize{\textcolor{red}{+2.1}} & 49.6\footnotesize{\textcolor{red}{+0.7}} & 42.1\footnotesize{\textcolor{red}{+2.7}} & 33.9\footnotesize{\textcolor{red}{+6.5}} & 52.8\footnotesize{\textcolor{red}{+1.9}} \\
        \checkmark & \checkmark & \checkmark & \textbf{51.3}\footnotesize{\textcolor{red}{+11.1}} & \textbf{56.6}\footnotesize{\textcolor{red}{+7.0}} & \textbf{58.2}\footnotesize{\textcolor{red}{+16.1}} & \textbf{50.5}\footnotesize{\textcolor{red}{+16.6}} & \textbf{68.0}\footnotesize{\textcolor{red}{+15.2}} \\
        \bottomrule
        \end{tabular}%
        }
    \end{minipage}
    \vspace{-4pt}
\end{table}

\subsection{Studies on Diffusion Detector}

\noindent \textbf{Components of {Diff. Detector}}: We present ablation studies in Tab.~\ref{tab:diff-detector}. \textit{Domain Augmentation}, \textit{Feature Collection}, and \textit{Feature Fusion} improve both source domain fitness and target domain generalization. The \textit{Auxiliary Branch} builds a more robust diffusion detector, while the \textit{Consistency Loss} enhances cross-domain generalization with minimal impact on source domain accuracy, validating its role in balancing fitness and generalization. The Inference Time results show \textit{Feature Collection} and \textit{Feature Fusion} mitigate accuracy drops from single-step diffusion efficiently. \textit{Domain Augmentation}, \textit{Auxiliary Branch}, and \textit{Consistency Loss} only apply during training with no impact on inference speed.

\noindent \textbf{Studies of Loss Weights $\gamma$ and $\lambda$}: As shown in Tab.~\ref{tab:loss_weights}, both $\gamma$ and $\lambda$ demonstrate stable performance within $[0.5, 1.5]$, because ${L}_{cons}$ gradually decreases during the later training steps, minimally affecting the final results.

\noindent \textbf{Results and Comparisons of {SD-2.1}}: We present results using SD-2.1 weights in Tab.~\ref{tab:sd-version}. Similar to SD-1.5, our method demonstrates consistent improvements across all settings compared to DDT~\cite{ddt} (average \textbf{12.3\%} on diff. detector and \textbf{4.1\%} on diff. guided DA) and GDD~\cite{gdd} (average \textbf{5.4\%} on diff. detector and \textbf{3.1\%} on diff. guided DG), confirming the generality of our approach.

\subsection{Diff. Guided Detector for DG and DA}

We present the Diff. guided detector's performance in DG and DA tasks (Tab.~\ref{tab:DG},~\ref{tab:DA}). \textit{Domain Augmentation} brings significant improvements (average \textbf{11.4\%}), while the diff. detector enhancement is relatively minor (average \textbf{3.3\%}, Tab.~\ref{tab:diff-detector}). With unlabeled target domain data, pseudo-label generation (\textit{Object Alignment}) is crucial, bringing average \textbf{13.1\% mAP}. Without target domain data, both \textit{Object Alignment} and \textit{Feature Alignment} are essential, contributing average \textbf{5.43\%} and \textbf{1.8\%} mAP respectively.

\subsection{{Limitations}}

Although our work improves diffusion detector inference efficiency, we do not implement engineering techniques like model distillation or TensorRT acceleration due to time constraints. Our method is only tested on official Stable-Diffusion weights, without exploring specialized instance-related models~\cite{instancediffusion,geodiffusion,detdiffusion} that might better fit detection tasks. Additionally, our experiments in Tab.~\ref{tab:voc} reveal that in DG tasks where target domain data is inaccessible, transferring generalization from diffusion detectors to ordinary detectors still faces limited improvement, indicating that more efficient transfer structures deserve further investigation.

\section{Conclusion}

In this paper, we present a framework for domain generalized and adaptive detection using diffusion models. Our approach optimizes single-step feature collection and fusion structures to reduce inference time by \textbf{75\%}, incorporates an object-centered auxiliary branch and consistency loss to enhance generalization for diff. detector. Then we transfer the generalization capabilities to standard detectors through feature- and object-level alignment strategies. Through comprehensive experiments across multiple benchmarks, we demonstrate average performance improvements of up to \textbf{3.0\%} (diff.~detector), \textbf{1.3\%} mAP (diff.~guided detector) for DG tasks and \textbf{5.4\%} mAP for DA. When scaled to larger datasets, our approach maintains substantial advantages, particularly in scenarios with large domain shifts and limited training data. This work offers practical solutions for generalized and adaptive detection and provides valuable insights for visual perception tasks requiring strong generalization and adaptation capabilities.

\clearpage
\clearpage
{
    \small
    \bibliographystyle{ieeenat_fullname}
    \bibliography{main}
}
\end{document}